\documentclass[lettersize,journal]{IEEEtran}
\usepackage{amsmath,amsfonts}
\usepackage{algorithmic}
\usepackage{algorithm}
\usepackage{array}
\usepackage[caption=false,font=normalsize,labelfont=sf,textfont=sf]{subfig}
\usepackage{booktabs}
\usepackage{multirow}
\usepackage{tabularx}
\usepackage{textcomp}
\usepackage{stfloats}
\usepackage{url}
\usepackage{verbatim}
\usepackage{graphicx}
\usepackage{cite}

\begin{document}

\title{Task-KV: Task-aware KV Cache Optimization via Semantic Differentiation of Attention Heads}

\author{Xingyang He, Jie Liu, Shaowei Chen
\thanks{Xingyang He is with the College of Artificial Intelligence, NanKai University, Tianjin 300350, China (e-mail: xingyanghe@mail.nankai.edu.cn).}
\thanks{Jie Liu is with the College of Artificial Intelligence, NanKai University,
Tianjin 300350, China (e-mail: jliu@nankai.edu.cn).}
\thanks{Shaowei Chen is with the College of Artificial Intelligence, NanKai University,
Tianjin 300350, China (e-mail: shaoweichen@mail.nankai.edu.cn).}
}



\maketitle

\begin{abstract}
KV cache is a widely used acceleration technique for large language models (LLMs) inference. However, its memory requirement grows rapidly with input length. 
Previous studies have reduced the size of KV cache by either removing the same number of unimportant tokens for all attention heads or by allocating differentiated KV cache budgets for pre-identified attention heads. 
However, due to the importance of attention heads varies across different tasks, the pre-identified attention heads fail to adapt effectively to various downstream tasks. To address this issue, we propose Task-KV, a method that leverages the semantic differentiation of attention heads to allocate differentiated KV cache budgets across various tasks. We demonstrate that attention heads far from the semantic center (called heterogeneous heads) make an significant contribution to task outputs and semantic understanding. In contrast, other attention heads play the role of aggregating important information and focusing reasoning. Task-KV allocates full KV cache budget to heterogeneous heads to preserve comprehensive semantic information, while reserving a small number of recent tokens and attention sinks for non-heterogeneous heads. Furthermore, we innovatively introduce middle activations to preserve key contextual information aggregated from non-heterogeneous heads. To dynamically perceive semantic differences among attention heads, we design a semantic separator to distinguish heterogeneous heads from non-heterogeneous ones based on their distances from the semantic center. Experimental results on multiple benchmarks and different model architectures demonstrate that Task-KV significantly outperforms existing baseline methods. Notably, in scenarios requiring full-context processing, such as summarization and synthetic tasks, Task-KV achieves performance comparable to the full KV cache while utilizing only 40\% of the memory.
\end{abstract}


\section{Introduction}
\IEEEPARstart{L}{LMs} are widely utilized in long-context scenarios such as in-context learning \cite{dong2022survey, qin2023tool}, multi-turn conversations \cite{fabbri2019multi,li2024streamingdialogue}, and retrieval-augmented \cite{gao2023retrieval,wang2024biorag} tasks. To improve inference speed and efficiency, LLMs reduce redundant computations by caching the Key and Value states (KV cache) of all tokens across all attention heads \cite{zhang2024h2o,lu2024longheads}. However, as the length of the input sequence increases, the storage requirement of KV cache expands dramatically, posing significant challenges to memory capacity and inference efficiency.

\begin{figure}[t]
    \begin{center}
		\includegraphics[width=0.9\linewidth]{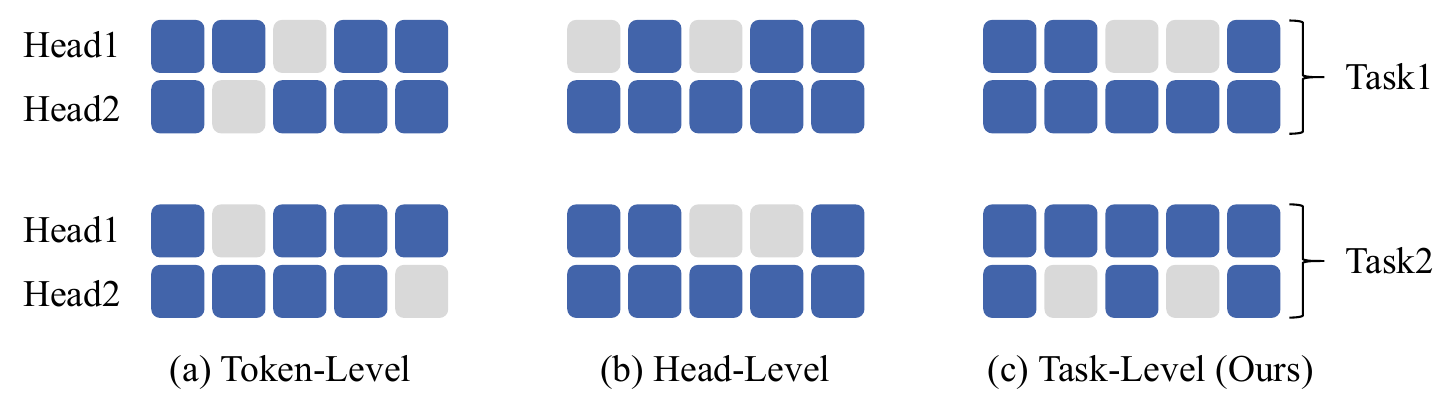}
    \end{center}
    \caption{Illustration of Task-KV compared with existing KV cache compression methods. (a) Token-level methods allocate the same KV cache budget to each attention head. (b) Head-level methods pre-identify important attention heads, but the KV cache budget among these heads remains fixed regardless of the task. (c) Our method identifies important attention heads based on the specific task and dynamically adjusts the KV cache budget among attention heads according to task semantics. }
    \label{difference_level}
\end{figure}

To address this issue, researchers have proposed various KV cache compression methods, mainly starting from two dimensions: token-level and head-level, as shown in Fig. \ref{difference_level}. Token-level \cite{xiao2023efficient,zhang2024pyramidkv,yang2024pyramidinfer,li2024snapkv, liu2024minicache} compression methods evict a fixed number of unimportant tokens for each attention head, aiming to reduce the KV cache size while preserving generation quality as much as possible. Head-level \cite{tang2407razorattention,feng2024ada,xiao2024duoattention,fu2024not} compression methods, on the other hand, pre-identify important attention heads \cite{wu2024retrieval,fu2024not} (e.g., retrieval heads) through experiments and allocate KV cache budgets based on their significance during inference, thereby further optimizing KV cache compression. However, the importance of attention heads varies across different tasks \cite{ma2021contributions,zhang2023closer}, meaning that the pre-identified important attention heads may not be universally critical for all tasks. A key challenge remains: how to adaptively identify and select critical attention heads based on task-specific requirements.

In this paper, we propose a novel approach called Task-KV, which dynamically allocates KV cache budgets by leveraging task-aware semantic differences among attention heads. Through theoretical analysis and empirical validation, we demonstrate that attention heads far from the semantic center, referred to as heterogeneous heads, make particularly significant contributions to task outputs. These heterogeneous heads capture the semantic information of the task from different perspectives, which is crucial for LLMs to fully understand the task semantics. The remaining non-heterogeneous heads, on the other hand, are mainly responsible for information aggregation and inference, and tend to process similar semantic information.
Based on these findings, Task-KV allocates the full KV cache budget to heterogeneous heads to preserve the completeness of multi-perspective semantic information. For non-heterogeneous heads, we retain only a small number of recent tokens and attention sinks to maintain basic inference capabilities. However, limiting storage to these tokens alone may lead to significant information gaps. To address this, we selectively retain a small subset of tokens with high attention scores from intermediate positions, referred to as middle activations, which effectively capture the critical contextual information aggregated by non-heterogeneous heads.
To dynamically perceive semantic differences among attention heads based on task requirements, we design a simple yet efficient semantic separator. This separator calculates the semantic vectors of attention heads by selecting task-relevant tokens and distinguishes heterogeneous heads from non-heterogeneous ones based on their distances from the semantic center.

We conduct extensive experiments across multiple benchmark tasks \cite{bai2023longbench,li2023loogle} and different model architectures \cite{ainslie2023gqa,touvron2023llama,jiang2023mistral} to validate the effectiveness of Task-KV. The results demonstrate that Task-KV significantly outperforms existing baseline methods in a variety of long-context tasks. Notably, in scenarios requiring processing of complete context, such as summarization and synthetic tasks, Task-KV achieves performance comparable to a full KV cache while utilizing only 40\% of the KV cache budget. 

In summary, our contributions are as follows:
\begin{itemize}
    \item We identify that attention heads far from the semantic center (heterogeneous heads) have a substantial impact on task outputs and validate this conclusion through both theoretical analysis and experimental evidence.

    \item We propose the Task-KV method, which dynamically distinguishes between heterogeneous and non-heterogeneous heads based on task-aware semantic differences among attention heads. By allocating differentiated KV cache budgets for different categories of attention heads, Task-KV effectively balances inference efficiency and generation quality.

    \item We demonstrate the superiority of Task-KV through comprehensive experiments on multiple benchmarks and different model architectures and conduct ablation studies to analyze the effectiveness of its individual components.
\end{itemize}

\section{Related work}
\subsection{Token-level KV compression methods}
Optimizing the KV cache has become a critical strategy for managing long sequences and reducing memory usage \cite{xiao2024infllm,lu2024longheads}. Prior research primarily focuses on selecting significant tokens and caching only their KV states to minimize KV cache size while maintaining model performance. For instance, 
Xiao et al. \cite{xiao2023efficient} retains only attention sinks and recent tokens, restoring the sliding window mechanism to handle long contexts effectively.
Li et al. \cite{li2024snapkv} enhances efficiency by compressing KV caches through the selection of significant KV positions based on attention scores.
Zhang et al. \cite{zhang2024h2o} employs a dynamic eviction policy that balances the retention of recent and historically significant tokens, optimizing memory usage while preserving essential information.
Liu et al. \cite{liu2024minicache} leverages similarities in KV caches across layers, enabling compression by caching KV states for only a subset of layers and reconstructing the states for other layers during decoding.
Zhang et al. \cite{zhang2024pyramidkv} allocates progressively reduced KV cache budgets across layers following a pyramid structure, further optimizing information transmission during KV cache compression.
However, these methods allocate the same KV cache budget to all attention heads, which may lead to the omission of crucial information in key attention heads. In contrast, our method assigns differentiated KV cache budgets to different types of attention heads, effectively reducing information loss during KV cache compression.

\subsection{Head-level KV compression methods}
Recent research has begun to explore head-level methods for compressing the KV cache. For example, Ge et al. \cite{ge2023model} applies various fixed compression strategies based on the characteristics of the attention head, but it relies on attention weights rather than semantic information.
Feng et al. \cite{feng2024ada} optimizes the Top-k selection algorithm by identifying important tokens from a global perspective, but it still risks overlooking critical attention heads.
Tang et al. \cite{tang2407razorattention}, Xiao et al. \cite{xiao2024duoattention}, and Fu et al. \cite{fu2024not} pre-identify important attention heads (e.g., retrieval or retrieval-reasoning heads) and allocate KV cache budgets according to their importance.
Although these methods are highly effective, they may not fully optimize the KV cache allocation for downstream tasks. In contrast, our method exhibits task-awareness by recognizing the semantic differences among attention heads and allocates KV cache budgets based on the specific semantic requirements of each task.

\section{Motivation}
\label{Motivation}
In this section, we first explore the semantic heterogeneity among attention heads and empirically demonstrate that heterogeneous heads are critical for maintaining model's performance (Section \ref{Heterogeneous Heads}). Next, we experimentally verify that there are significant differences in heterogeneous heads activated by different tasks (Section \ref{Task variability}). Finally, we provide a theoretical analysis, establishing that heterogeneous heads are the key factor determining the upper bound of the LLMs' output contribution, further highlighting their pivotal role in LLMs inference (Section \ref{Theoretical analysis}).

\begin{figure*}[ht]
    \begin{center}
		\includegraphics[width=0.85\linewidth]{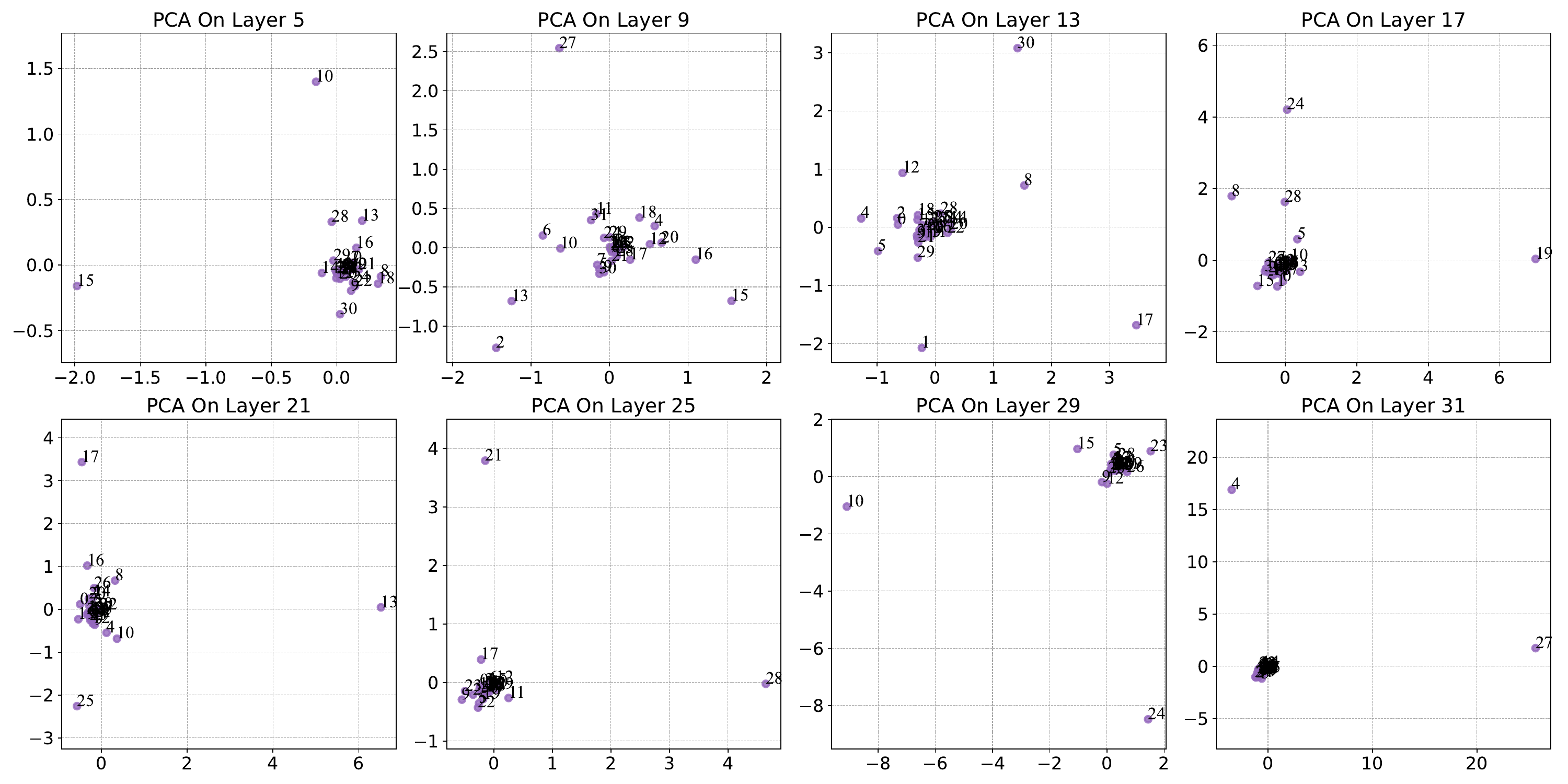}
    \end{center}
    \caption{For each specific layer, we use PCA to reduce the semantic vectors of different attention heads to two dimensions for visualization, allowing us to observe the differences between the semantic vectors. }
    \label{pca}
\end{figure*}

\begin{figure*}[ht]
    \begin{center}
		\includegraphics[width=0.85\linewidth]{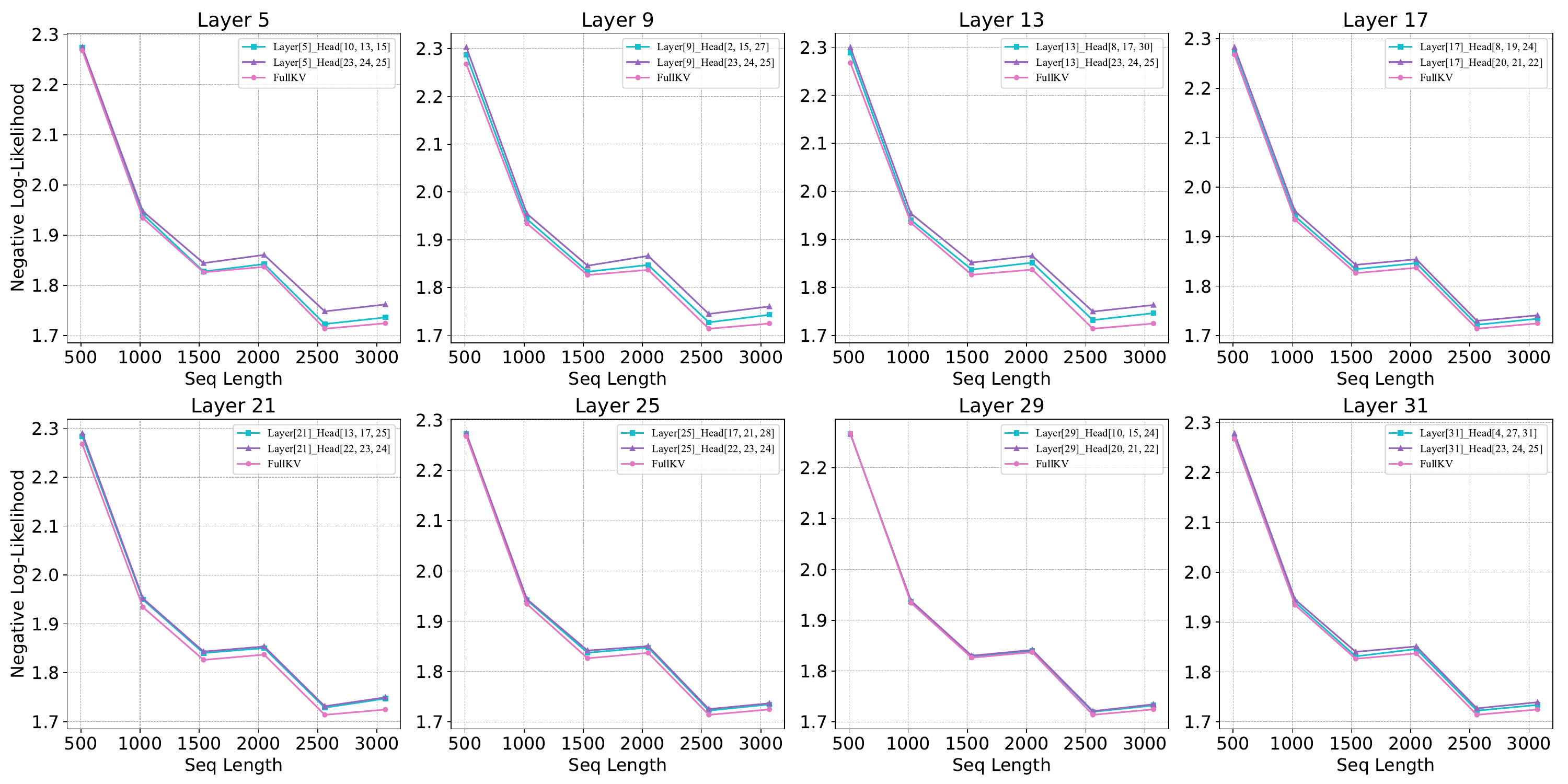}
    \end{center}
    \caption{Results of comparative experiments in which only heterogeneous or non-heterogeneous heads were retained in different layers. }
    \label{post-nll}
\end{figure*}

\subsection{Heterogeneous Heads}
\label{Heterogeneous Heads}
Although $Q$, $K$, and $V$ in the attention mechanism all contain semantic information, the attention mechanism itself functions as a weighted average of $V$, meaning that only the semantic information in $V$ is propagated to the next layer and contributes to the model's output. Consequently, we focus on analyzing the $V$ in the attention heads to investigate how the semantic differences between attention heads influence the model's output. Specifically, for each attention head, after computing the attention weight matrix $A$, we average by columns to obtain the weight distribution of the attention head to the current context, and then weighted sum over $V$ to derive the semantic vector of the attention head. The formula for this process is as follows: 
\begin{align}
    A &= Softmax(QK^{T}/\sqrt{d}  +M) \\
    v &=\frac{\sum_{i=1}^{N}A[i,:] }{N} \cdot V
    \label{eq:semantic vector}
\end{align}
where $Q,K,V\in \mathbb{R} ^{N\times d} $ are query states, key states, value states respectively, $M \in \mathbb{R}^{N\times N}$ is the mask matrix, $v \in \mathbb{R} ^{1\times d} $ is the semantic vector that highly summarizes the semantic information the current attention head is focusing on.

To more intuitively observe the semantic differences between different attention heads, we apply Principal Component Analysis (PCA) \cite{mackiewicz1993principal} to downscale the semantic vectors of the attention heads to two dimensions and perform visualization analysis. As shown in Fig. \ref{pca}, within the semantic space of each layer, most attention heads cluster closely together, while a smaller subset is positioned farther from the semantic center. We hypothesize that these attention heads, distant from the semantic center, encode semantic information from diverse perspectives, and they are essential for the model’s comprehensive understanding of task semantics. We refer to them as heterogeneous heads. To validate this hypothesis, we select three attention heads from each of the heterogeneous and non-heterogeneous heads for the control experiment. Specifically, for layer 9, we retain only the selected three heterogeneous heads or three non-heterogeneous heads while removing the remaining attention heads in that layer. The attention heads in all other layers are kept unchanged. We then calculate the negative log-likelihood (NLL) to measure the divergence between the outputs and the standard model's outputs. As shown in Fig. \ref{post-nll}, the NLL curve is closer to the standard output when the heterogeneous heads are retained, which better preserves the model's original performance compared to retaining the non-heterogeneous heads.

The reason behind this phenomenon is clear: the semantic distinctiveness of the heterogeneous heads enhances the expressive and generalization capabilities of the model.  Retaining these heads can take full advantage of the diverse semantic information extracted by the multi-head attention (MHA) mechanism \cite{vaswani2017attention}, which is the original purpose of the design of the MHA. In contrast, the semantic information of non-heterogeneous heads is more homogeneous. Retaining only non-heterogeneous heads results in information loss, leading to greater deviation between the outputs and those of the standard model. However, this does not mean that non-heterogeneous heads are useless. They mainly serve the function of information aggregation and reasoning. Their absence can seriously affect the inference ability of the model. We have a detailed discussion in Section \ref{Effect of non-heterogeneous heads}.

\subsection{Task variability among attentional heads}
\label{Task variability}
\begin{figure*}[ht]
    \begin{center}
		\includegraphics[width=0.8\linewidth]{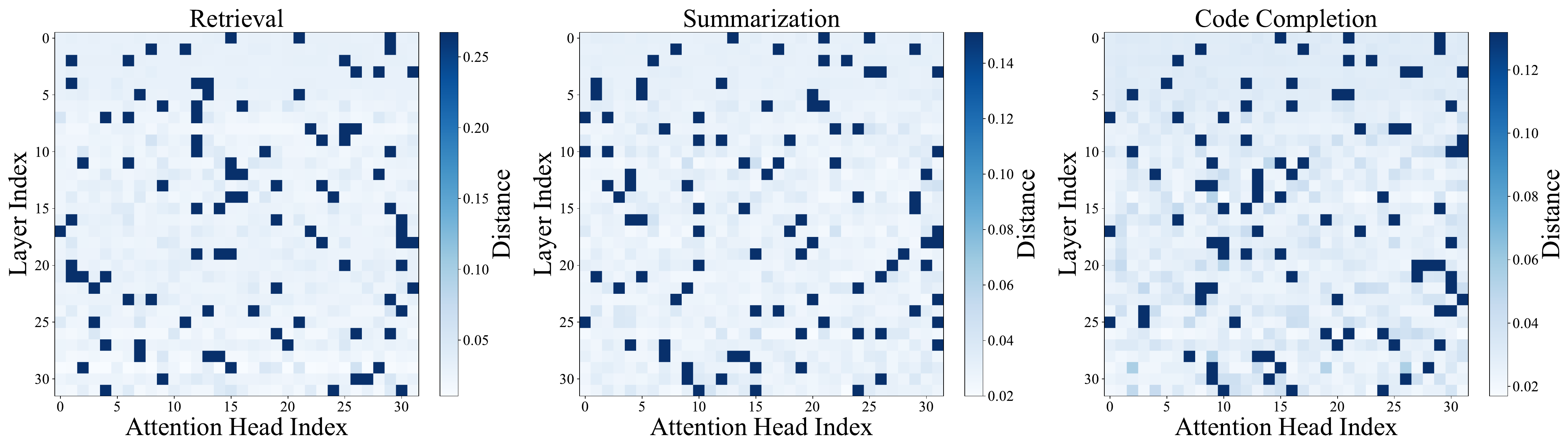}
    \end{center}
    \caption{Distribution of heterogeneous heads across different tasks within the Llama-2-7B-Chat model}
    \label{llama_distance}
\end{figure*}

\begin{figure*}[ht]
    \begin{center}
		\includegraphics[width=0.4\linewidth]{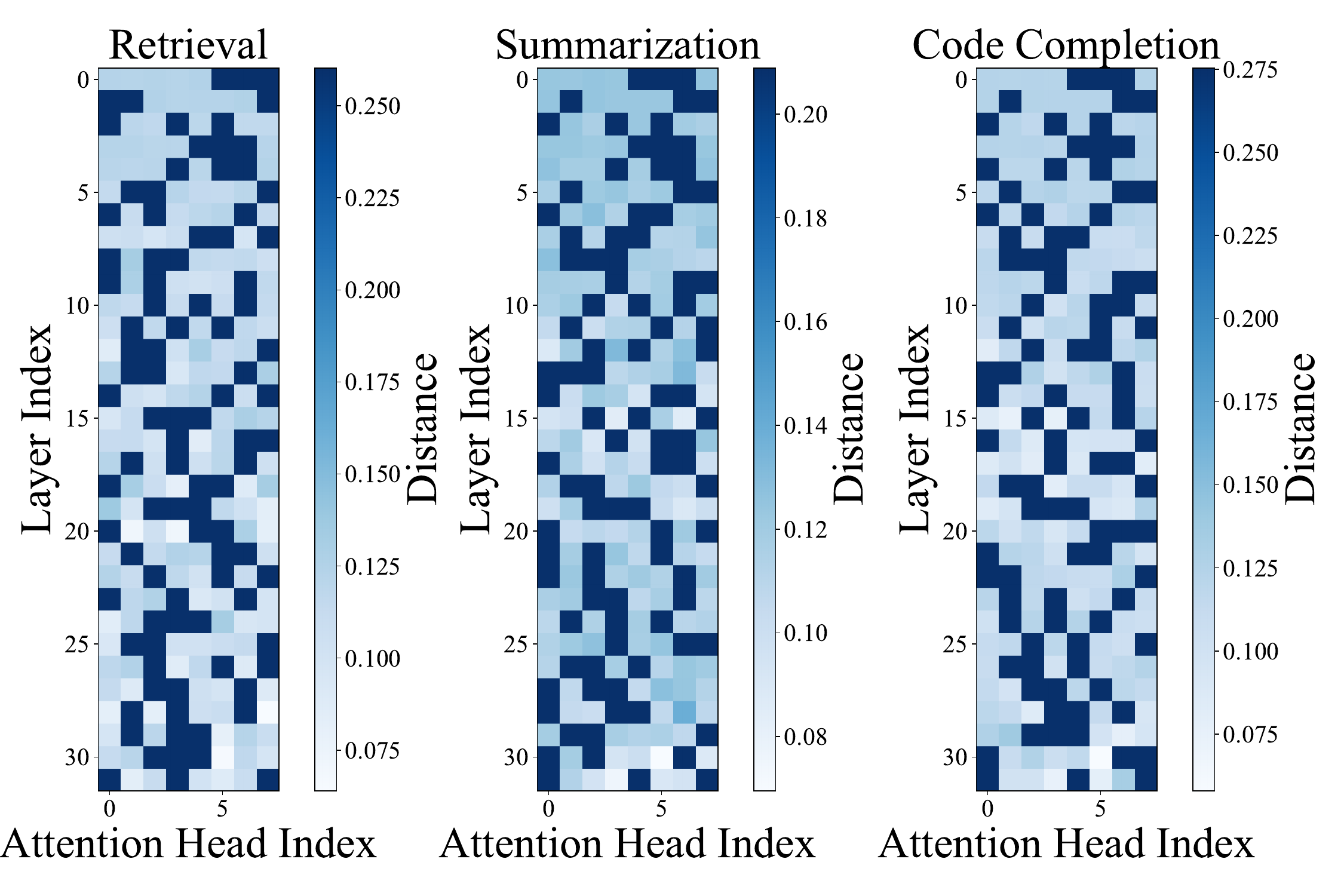}
    \end{center}
    \caption{Distribution of heterogeneous heads across different tasks within the Mistral-7B-v0.2-Instruct model }
    \label{mistral_distance}
\end{figure*}

To further investigate the distribution characteristics of heterogeneous heads in different tasks, we select three attention heads furthest from the semantic center in each layer for visualization across different tasks. As shown in Fig. \ref{llama_distance}, \ref{mistral_distance}, we analyze the distribution characteristics of activated heterogeneous heads in retrieval, summarization, and code completion tasks across different model architectures. It can be clearly seen that the distribution of heterogeneous heads is directly influenced by the task objectives. The semantic representation requirements of different tasks determine the distribution characteristics of heterogeneous heads in the semantic space.

\subsection{Theoretical analysis}
\label{Theoretical analysis}
The contribution of a particular attention head to the model's output can be interpreted as the degree of change in the model's output after removing that attention head. Therefore, the contribution of the $j$-th attention head to the model's output $y$ can be defined as:
\begin{align}
    \left \| \Delta y_{j} \right \| ^{2}=\left \| y-y_{\mathcal{H } \setminus \{h_j\}}  \right \|^{2} 
\end{align}
where $\mathcal{H }$ denotes the set of attention heads, $h_{j}$ denotes the $j$-th attention head, $\mathcal{H } \setminus \{h_j\} $ denotes the removal of $h_{j}$ from $\mathcal{H}$, $\left \| \cdot  \right \| $ denotes the $L_{2}$ norm.

To simplify the analysis, we study the case of one layer of LLM and consider only the output of the MHA. The output of MHA is obtained by splicing the outputs of different attention heads and then transforming them linearly, so $y$ and $y_{\mathcal{H } \setminus \{h_j\}} $ can be expressed as:
\begin{align}
    y&=\left [ v_{1},...,v_{n}   \right ]W_{o} \nonumber \\
    &=\sum_{1 \leq i \leq n} v_{i} W_{o,i} 
\end{align}
\begin{align}
    y_{\mathcal{H } \setminus \{h_j\}} =\sum_{1 \leq i \leq n, i \neq j} v_{i} W_{o,i} 
\end{align}
where $v_{i} \in \mathbb{R} ^{N\times d}$ is the result of the calculation of the $i$-th attention head, $W_{o} \in \mathbb{R} ^{nd\times N}$ is a linear projection layer, $W_{o_{i}} \in \mathbb{R} ^{d\times N}$ is the $i$-th block matrix of $w_{o}$, $n$ is the number of attention heads.

Therefore, $\left \| \Delta y_{j} \right \| ^{2}$ can be formulated as:
\begin{align}
    \left \| \Delta y_{j} \right \| ^{2}=\left \| v_{j}\cdot W_{o,j}   \right \| ^{2}  
    \label{eq:contribution}
\end{align}

Denote $v_{j}$ as the sum of the mean vector $\widetilde{v} $ and the offset $\delta _{j} $:
\begin{align}
    v_{j}&=\widetilde{v} + \delta _{j} \\
    \widetilde{v}&=\frac{\sum_{i=1}^{n} v_{i}  }{n} 
\end{align}

Substituting into Equation \eqref{eq:contribution} expands it:
\begin{align}
\left \| \Delta y_{j} \right \| ^{2}&=\left \| (\widetilde{v}+\delta _{j}  ) \cdot W_{o,j} \right \|^{2}  \nonumber \\
&=\left \| \widetilde{v} \cdot W_{o,j}  \right \|  ^{2}  +\left \| \delta _{j}\cdot W_{o,j}  \right \|^{2}+2\left \langle \widetilde{v}\cdot W_{o,j} , \delta _{j}\cdot W_{o,j} \right \rangle 
\end{align}

In practice, the elements in $W_{o,j}$ are usually a finite number, so it can be assumed to be a bounded matrix. Therefore,  $W_{o,j}$ satisfies:
\begin{align}
    \left \| W_{o,j} \right \| \le C,\forall j\in \left \{ 1,2,...,n \right \} 
\end{align}
where $C$ is a bounded constant.

So $\left \| \Delta y_{j} \right \| ^{2}$ can be deflated as:
\begin{align}
\left \| \Delta y_{j} \right \| ^{2}\le \left ( \left \| \widetilde{v}  \right \|^{2} +\left \| \delta _{j}  \right \| ^{2} +2\left \| \widetilde{v}  \right \| \left \| \delta _{j}  \right \|    \right )C^{2}
\label{eq:bounded}
\end{align}

As shown in Equation \eqref{eq:bounded}, for different attention heads, both $\widetilde{v}$ and $C$ remain constant. The factor that truly influences the upper bound of the contribution to the model's output is the offset $\delta _{j} $. This observation provides a theoretical explanation for the higher contribution of heterogeneous heads to the model's output.

\begin{figure*}[!t]
    \begin{center}
		\includegraphics[width=1\linewidth]{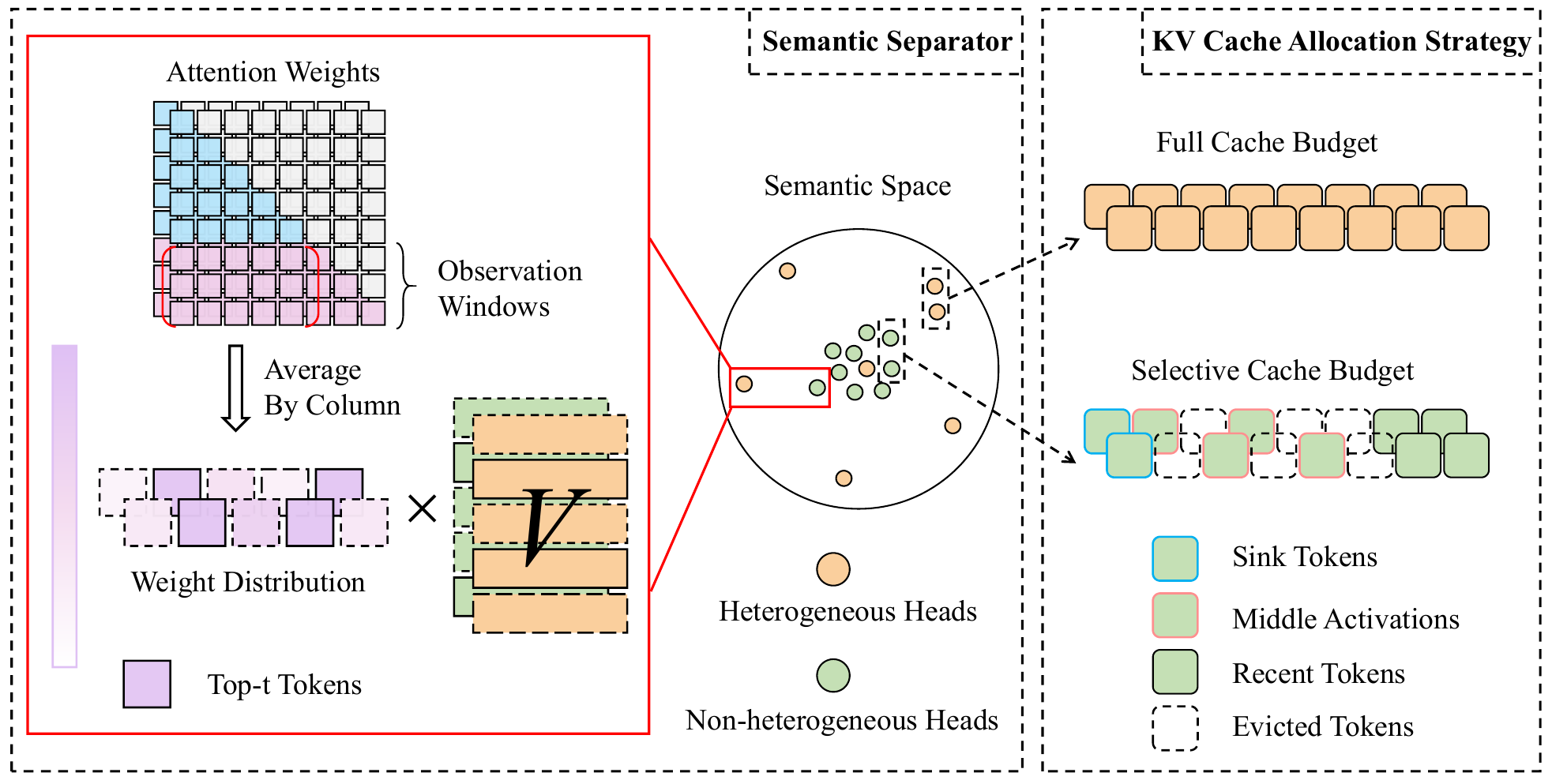}
    \end{center}
    \caption{Overview of Task-KV. }
    \label{Task-KV}
\end{figure*}

\section{Task-KV}
Motivated by the above insights, we propose a novel method called Task-KV, designed to dynamically allocate KV cache budgets by leveraging task-aware semantic differences among attention heads. As illustrated in Fig. \ref{Task-KV}, Task-KV comprises two key components: (1) a semantic separator, which efficiently and accurately distinguishes heterogeneous heads from non-heterogeneous heads based on their semantic differences (Section \ref{Semantic separator}); and (2) a KV cache allocation strategy, which allocates differentiated KV cache budgets to different types of attention heads and determines the critical KV states to retain for each head (Section \ref{KV Cache allocation strategy}).

\subsection{Semantic separator}
\label{Semantic separator}
Normally, the semantic vectors of attention heads should be computed according to Equation (1)(2). However, calculating the complete attention weight matrix introduces significant computational costs, which is detrimental to inference acceleration. Inspired by Li et al. \cite{li2024snapkv}, we adopt a more efficient approach by using only a small portion of the segment at the end of the input sequence as the observation window. This allows us to compute a local weight matrix to approximate the semantic information. The local weight matrix $A^{\prime  }\in \mathbb{R}^{L\times N}$ is calculated as follows:
\begin{align}
    A^{\prime  } &= Softmax(Q[-L:,:]\cdot K^{T}/\sqrt{d}  +M^{\prime  }) 
\end{align}
where $L$ denotes the observation window size, $M^{\prime  } \in \mathbb{R}^{L\times L}$ is the mask matrix.

Subsequently, we average the weight matrices by columns and compute a weighted sum with the corresponding value states to generate the semantic vectors for each attention head. However, the computational costs incurred by this process remains unacceptable when the input sequence is long. Based on previous studies \cite{zhang2024pyramidkv, xiao2023efficient, han2023lm}, a small number of tokens often account for the majority of attention scores, we select only the top $t$ tokens with the highest attention scores to compute the semantic vectors. This approach significantly reduces computational costs while maintaining results comparable to those obtained using the full sequence. The specific formula is as follows: 
\begin{align}
    C&=\frac{\sum_{i=1}^{L} A^{\prime }\left [ i,: \right ] }{L} \\
    I&=Top_{k} ( C,t) \\
    v^{\prime}& =C[I,:]\cdot V[I,:]
\end{align}
where $I$ denotes the index of the top $t$ score selected from $C$, $v^{\prime}$ is the semantic vector of the current attention head. With these two optimization steps, we significantly reduce the computational costs of semantic vectors.

\begin{table*}[t]
\renewcommand{\arraystretch}{1}
\setlength{\tabcolsep}{3pt}
\centering
\caption{Details of LongBench and LooGLE.}
\scalebox{1}{
\begin{tabular}{l|llcrc|c}
\toprule
Source  & Task & Task Type & Eval metric & Avg Len & Language & Nums \\
\midrule
\multirow{12}{*}{LongBench}  & Qasper & Single-Doc. QA & F1 & 3,619 & EN & 200 \\
  & MultiFieldQA-en & Single-Doc. QA & F1 & 4,559 & EN & 150 \\
 & HotpotQA & Multi-Doc. QA & F1 & 9,151 & EN & 200 \\
 & 2WikiMultihopQA & Multi-Doc. QA & F1 & 4,887 & EN & 200 \\
 & GovReport & Summarization & Rouge-L & 8,734 & EN & 200 \\
  & QMSum & Summarization & Rouge-L & 10,614 & EN & 200 \\
  & TREC & Few-shot Learning & Accuracy & 5,177 & EN & 200 \\
  & TriviaQA & Few-shot Learning & F1 & 8,209 & EN & 200 \\
  & PassageCount & Synthetic Task & Accuracy & 11,141 & EN & 200 \\
  & PassageRetrieval-en & Synthetic Task & Accuracy & 9,289 & EN & 200 \\
  & LCC & Code Completion & Edit Sim & 1,235 & Python/C++/Java & 200 \\
  & RepoBench-P & Code Completion & Edit Sim & 4,206 & Python/Java & 500 \\
\midrule
\multirow{3}{*}{LonGLE}  & Computation & Long-Dep. QA & F1 & 17,001 & EN & 100 \\
 & Multiple Information Retrieval & Long-Dep. QA & F1 & 14,808 & EN & 100 \\
  & Long Dependency Summarization & Long-Dep. Sum. & Rouge-L & 20,887 & EN & 100 \\
\bottomrule
\end{tabular}
}
\label{Details of Datasets}
\end{table*}

Next, we rank the attention heads based on the distance from the semantic center, selecting a certain number of heads from farthest to nearest as heterogeneous heads. The remaining heads are classified as non-heterogeneous. While the heterogeneous heads capture diverse semantic features, they lack the aggregated semantic information typically provided by the non-heterogeneous heads. To address this, we select the attention head closest to the semantic center from the non-heterogeneous set and incorporate it into the heterogeneous head set. This ensures that the heterogeneous heads can cover all types of semantic information.

Moreover, as observed in Fig. \ref{pca}, the number of heterogeneous heads decreases progressively across layers as the model depth increases. To accommodate this trend, we select a larger number of heterogeneous heads in lower layers and fewer in higher layers. Specifically, we define the parameter $\beta $ to represent the proportion of heterogeneous heads in the bottom layer and the parameter $m$ to denote the number of heterogeneous heads in the top layer. For intermediate layers, the number of heterogeneous heads is determined through linear interpolation. The number of heterogeneous heads in the $r$-th layer is:
\begin{align}
    f(r)=n\beta -\frac{n\beta -m}{R-1} \cdot r, \ r=0,1,...,R-1
\end{align}
where $n$ is the number of attention heads and $R$ is the number of transformer layers of the model.

\subsection{KV Cache allocation strategy}
\label{KV Cache allocation strategy}
For heterogeneous heads, we allocate the full KV cache budget to ensure the completeness of diverse semantic information. For non-heterogeneous heads, we adopt a selective retention strategy by preserving a small number of the most recent tokens and attention sinks to maintain basic inference capabilities. Additionally, we select a small set of tokens with the highest attention scores from the intermediate portion of the sequence. These tokens, referred to as middle activations, aggregate critical contextual information and provide precise guidance for model inference (a detailed analysis is presented in Section \ref{Importance of middle activations}). The number of middle activations k, is determined by the following formula:
\begin{align}
    k = \frac{B-N\cdot f(r)}{n-f(r)} -s_{1}-s_{2} 
\end{align}
where $B$ denotes the total KV cache budget of the current layer, $N$ denotes the sequence length, $s_{1}$ denotes the number of sink tokens, $s_{2}$ denotes the number of recent tokens.

\section{Experiment}
In this section, we first introduce the baselines (Section \ref{Baselines}), evaluation datasets (Section \ref{Datasets}), and backbone LLMs (Section \ref{Backbone LLMs}), followed by a detailed description of the experimental setup for Task-KV (Section \ref{setup}). Finally, we compare the performance of Task-KV with the baselines in the following three aspects: (1) a comprehensive evaluation of the model's ability to handle various long-context tasks (Section \ref{Long-context understanding tasks}); (2) an assessment of its performance in long-context retrieval and reasoning using the Reasoning-in-a-Haystack task \cite{fu2024not} (Section \ref{Reasoning-in-a-Haystack}); and (3) an evaluation of the model's memory footprint and computational efficiency in long-context scenarios (Section \ref{Memory and latency}).

\subsection{Baselines}
\label{Baselines}
We select StreamingLLM \cite{xiao2023efficient} as the KV cache compression method based on attention sinks, while SnapKV \cite{li2024snapkv} and PyramidKV \cite{zhang2024pyramidkv} are chosen as baselines for token-level KV cache compression. Additionally, HeadKV \cite{fu2024not} is used as the baseline for head-level KV cache compression. By comparing compression techniques across these different levels, we aim to provide a more comprehensive evaluation of the effectiveness of each approach.

\subsection{Datasets}
\label{Datasets}
We choose two benchmarks for comprehensively evaluating the model's capabilities on various long context tasks: LongBench \cite{bai2023longbench} and LooGLE \cite{li2023loogle}. LongBench covers multiple types of long-context tasks, including single-document QA \cite{dasigi2021dataset}, multi-document QA \cite{yang2018hotpotqa, ho2020constructing}, summarization \cite{zhong2021qmsum,huang2021efficient}, few-shot learning \cite{joshi2017triviaqa,li2002learning}, synthetic tasks \cite{li2012exploiting} and code completion \cite{liu2023repobench,guo2023longcoder}. LooGLE \cite{li2023loogle} covers a variety of long dependency tasks, and we choose computation, multiple information retrieval, and long dependency summarization tasks to complement LongBench. The details of LongBench and LooGLE are shown in Table \ref{Details of Datasets}.

\begin{table*}[t]
\renewcommand{\arraystretch}{1}
\setlength{\tabcolsep}{3pt}
\centering
\caption{ Performance comparison on the LongBench and LooGLE benchmarks for Llama-2-7B-Chat and Mistral-7B-v0.2-Instruct.}
\scalebox{1}{
\begin{tabular}{l|ccccccc|cccc}
\toprule
\multirow{3}{*}{Method} & \multicolumn{7}{c|}{LongBench} & \multicolumn{4}{c}{LoogGLE} \\
\cmidrule(lr){2-8} \cmidrule(l){9-12}
& Single-Doc & Multi-Doc & \multirow{2}{*}{Summarization} & Few-shot & Synthetic & Code & \multirow{2}{*}{Avg.} & \multirow{2}{*}{Computation} & Multi-Info & Long-Dep & \multirow{2}{*}{Avg.} \\
& QA & QA & & Learning & Tasks & Completion & & & Retrieval & Sum. & \\
\midrule
\multicolumn{12}{c}{Llama-2-7B-Chat, KV Cache Budget=100\%} \\
\midrule
FullKV & 27.25 & 28.69 & 22.71 & 73.76 & 6.00 & 54.30 & 35.45 & 10.88 & 11.38 & 2.76 & 8.34 \\
\midrule
\multicolumn{12}{c}{Llama-2-7B-Chat, KV Cache Budget=40\%} \\
\midrule
StreamingLLM & 19.16 & 23.40 & 18.53 & 71.97 & 2.25 & 53.74 & 31.51 & 8.19 & 9.46 & 2.37 & 6.67 \\
SnapKV       & 26.99 & 28.99 & 21.20 & 73.59 & 5.75 & 54.16 & 35.11 & 9.58 & 10.87 & 2.36 & 7.60 \\
PyramidKV    & 27.70 & 28.24 & 21.29 & 73.67 & 5.50 & 54.09 & 35.08 & 9.84 & 11.04 & 2.41 & 7.76 \\
HeadKV-R2    & \textbf{27.77} & 29.03 & 21.34 & 73.58 & 5.25 & \textbf{54.22} & 35.20 & 10.01 & 10.96 & 2.44 & 7.80 \\
Task-KV      & 26.70 & \textbf{29.08} & \textbf{21.35} & \textbf{73.92} & \textbf{6.75} & 54.12 & \textbf{35.32} & \textbf{10.23} & \textbf{11.12} & \textbf{2.48} & \textbf{7.94} \\
\midrule
\multicolumn{12}{c}{Llama-2-7B-Chat, KV Cache Budget=60\%} \\
\midrule
StreamingLLM & 22.71 & 26.43 & 19.62 & 73.80 & 2.75 & 53.75 & 33.17& 8.39 & 9.31 & 2.44 & 6.71 \\
SnapKV       & 27.36 & 28.41 & 22.01 & 73.59 & 6.00 & 54.13 & 35.25 & 10.73 & 10.82 & \textbf{2.59} & 8.05 \\
PyramidKV    & 27.76 & 28.42 & 22.05 & 73.76 & 5.25 & \textbf{54.22} & 35.24 & 10.36 & 10.89 & 2.43 & 7.89 \\
HeadKV-R2    & \textbf{27.78} & 28.86 & 22.13 & \textbf{73.83} & 5.00 & 54.11 & 35.29 & 10.54 & 10.81 & 2.48 & 7.94 \\
Task-KV      & 26.78 & \textbf{28.91} & \textbf{22.27} & 73.71 & \textbf{6.25} & 54.20 & \textbf{35.35} & \textbf{10.78} & \textbf{11.02} & \textbf{2.59} & \textbf{8.13} \\
\bottomrule

\multicolumn{12}{c}{Mistral-7B-v0.2-Instruct, KV Cache Budget=100\%} \\
\midrule
FullKV & 41.14 & 35.56 & 27.33 & 78.62 & 47.45 & 48.71 & 46.47 & 12.25 & 18.22 & 3.23 & 11.23 \\
\midrule
\multicolumn{12}{c}{Mistral-7B-v0.2-Instruct, KV Cache Budget=40\%} \\
\midrule
StreamingLLM & 29.09 & 31.17 & 23.89 & 76.83 & 26.59 & 47.29 & 39.14 & 9.92 & 15.02 & 3.17 & 9.37 \\
SnapKV       & 40.89 & \textbf{35.02} & 25.92 & 77.56 & 47.20 & \textbf{48.71} & 45.88 & 9.89 & 17.78 & 3.25 & 10.31 \\
PyramidKV    & 40.06 & 34.32 & 25.81 & \textbf{78.37} & \textbf{47.45} & 48.25 & 45.71 & 10.64 & 17.35 & 3.35 & 10.45 \\
HeadKV-R2    & \textbf{40.91} & 34.96 & 25.76 & 78.24 & 47.32 & 48.44 & 45.94 & 10.66 & \textbf{17.82} & 3.41 & 10.63 \\
Task-KV      & 40.73 & 34.89 & \textbf{26.01} & 78.33 & \textbf{47.45} & 48.49 & \textbf{45.98} & \textbf{10.94} & 17.74 & \textbf{3.50} & \textbf{10.73} \\
\midrule
\multicolumn{12}{c}{Mistral-7B-v0.2-Instruct, KV Cache Budget=60\%} \\
\midrule
StreamingLLM & 32.81 & 32.64 & 25.08 & 77.87 & 31.46 & 47.61 & 41.24 & 9.70 & 16.03 & 3.13 & 9.62 \\
SnapKV       & 40.69 & 34.90 & 26.83 & 78.28 & 47.20 & \textbf{48.69} & 46.10 & 10.80 & 17.90 & 3.18 & 10.63 \\
PyramidKV    & 41.11 & \textbf{35.45} & 27.39 & 78.28 & 47.12 & 48.63 & 46.33 & 11.56 & 17.64 & 3.10 & 10.77 \\
HeadKV-R2    & 41.24 & 35.36 & 27.25 & 78.39 & 47.15 & 48.65 & 46.34 & 11.65 & 17.81 & \textbf{3.20} & 10.89 \\
Task-KV      & \textbf{41.31} & 35.12 & \textbf{27.44} & \textbf{78.54} & \textbf{47.45} & 48.65 & \textbf{46.42} & \textbf{11.78} & \textbf{18.14} & 3.14 & \textbf{11.02 }\\
\bottomrule
\end{tabular}
}
\label{longbench and Loogle} 
\end{table*}

\subsection{Backbone LLMs}
\label{Backbone LLMs}
In this experiment, we utilize two distinct types of open-source LLMs: Llama-2-7B-Chat \cite{touvron2023llama} and Mistral-7B-v0.2-Instruct \cite{jiang2023mistral}, to comprehensively compare the performance differences between Task-KV and baselines. Llama-2-7B-Chat employs a MHA mechanism, where Q, K, V have a one-to-one correspondence. In contrast, Mistral-7B-v0.2-Instruct model adopts a grouped query attention (GQA) mechanism \cite{ainslie2023gqa}, where each group of KV pairs can correspond to multiple queries.

\subsection{Experiment setup}
\label{setup}
We set $t=256$ for computing semantic vectors. For heterogeneous heads, we configure $\beta=0.25$, $m=4$ for Llama-2-7B-Chat, and $\beta=0.3$, $m=1$ for Mistral-7B-v0.2-Instruct. For non-heterogeneous heads, we set the number of sink tokens to 16 and recent tokens to 256 for both models. To ensure a fair comparison, we follow Zhang et al. \cite{zhang2024pyramidkv} and set an observation window size of 32 and an average pooling kernel size of 7 across all baselines and our method.

\subsection{Main results}
\subsubsection{Long-context understanding tasks}
\label{Long-context understanding tasks}
In Table \ref{longbench and Loogle}, we present a comprehensive evaluation of various long-context tasks from the LongBench and LooGLE benchmarks, comparing the performance under two resource-constrained scenarios: KV cache budgets of 40\% and 60\%. These scenarios represent different levels of resource limitations. Experimental results demonstrate that our method significantly outperforms all baselines in terms of average scores on both benchmarks.

Efficient allocation of resources is critical under resource-constrained conditions. StreamingLLM \cite{xiao2023efficient} retains only attention sinks and recent tokens, resulting in substantial information loss. PyramidKV \cite{zhang2024pyramidkv} and SnapKV \cite{li2024snapkv} allocate identical KV cache budgets to all attention heads, potentially overlooking critical information while retaining irrelevant or redundant content. HeadKV \cite{fu2024not}, which relies on pre-identified key attention heads, exhibits limitations in adapting to the diversity of long-context tasks. In contrast, our approach adaptively allocates differentiated KV cache budgets to various types of attention heads based on task characteristics. This enables it to efficiently handle diverse long-context tasks even in resource-constrained scenarios.

\begin{figure*}[ht]
    \begin{center}
		\includegraphics[width=0.85\linewidth]{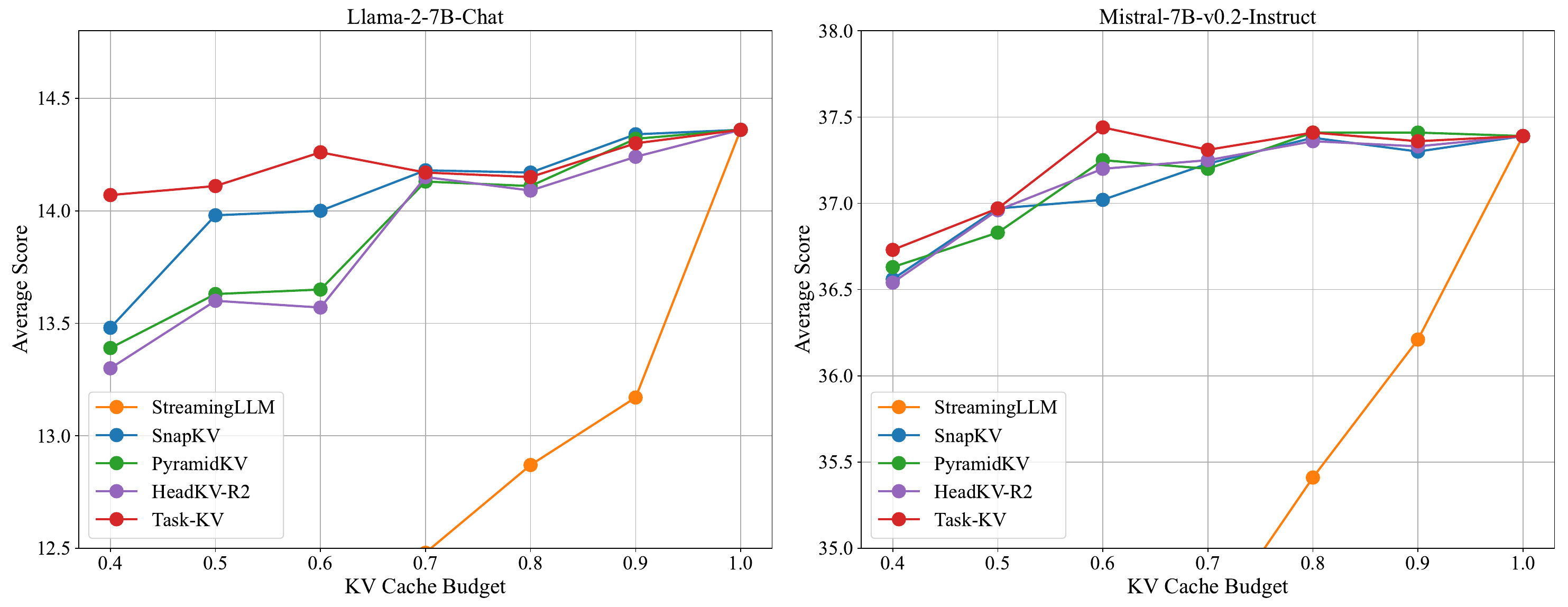}
    \end{center}
    \caption{Experimental results on summarization tasks and synthetic tasks under different KV cache budget conditions. The final experimental results are the average score of the two tasks }
    \label{summ_and_syn}
\end{figure*}

\begin{table*}[t]
\renewcommand{\arraystretch}{1.2}
\centering
\caption{Reasoning-in-a-Haystack test results with KV cache budget = 50\%}
\scalebox{1}{
\begin{tabular}{l|*{5}{>{\centering\arraybackslash}p{0.85cm}}|*{8}{>{\centering\arraybackslash}p{0.65cm}}}
\toprule
\multirow{2}{*}{Method} & \multicolumn{5}{c|}{Llama-2-7B-Chat, KV Cache Budget=50\%} & \multicolumn{8}{c}{Mistral-7B-v0.2-Instruct, KV Cache Budget=50\%} \\
\cmidrule(){2-14}
& 0k & 1k & 2k & 4k & Avg. & 0k & 1k & 2k & 4k & 8k & 16k & 32k & Avg. \\
\midrule
FullKV & 35.40 & 37.20 & 41.80 & 34.40 & 37.20 & 59.60 & 50.80 & 44.80 & 37.80 & 34.40 & 27.60 & 29.80 & 40.69 \\
\midrule
StreamingLLM  & \textbf{36.00} & 35.80 & 31.80 & 29.20 & 33.20 & 50.60 & 47.60 & 39.20 & 33.40 & 29.60 & 25.80 & 26.80 & 37.43 \\
SnapKV & 35.60 & 36.60 & \textbf{41.20} & 33.20 & 36.65 & \textbf{59.60} & 50.60 & 44.80 & 38.20 & 34.60 & 27.20 & \textbf{30.80} & 40.83 \\
PyramidKV  & 35.60 & 37.40 & 40.40 & 33.20 & 36.65 & 59.40 & 50.60 & 45.00 & \textbf{38.60} & \textbf{34.60} & 27.20 & 29.60 & 40.71 \\
HeadKV-R2  & \textbf{36.00} & 36.80 & \textbf{41.20} & 33.40 & 36.85 &\textbf{59.60} & 50.80 & 45.00 & 38.20 & 34.40 & 27.20 & 30.20 & 40.77 \\
Task-KV  & 35.20 & \textbf{38.20} & \textbf{41.20} & \textbf{34.00} & \textbf{37.15} & 59.40 & \textbf{51.20} & \textbf{45.40} & 38.40 & 34.40 & \textbf{27.40} & 30.40 & \textbf{40.94} \\
\bottomrule
\end{tabular}
}
\label{tab:Reasoning-in-a-Haystack}
\end{table*}
Notably, Task-KV exhibits superior performance in task scenarios that require spanning the complete context (e.g., summarization tasks and synthetic tasks). As illustrated in Fig. \ref{summ_and_syn}, we evaluate the Llama-2-7B-Chat and Mistral-7B-v0.2-Instruct models under varying KV cache budget conditions for summarization and synthetic tasks. In resource-constrained settings, Task-KV significantly outperformed existing baselines. This superiority is primarily attributed to the fact that Task-KV allocates a full KV cache budget for heterogeneous heads, enabling a comprehensive understanding of global semantic information.

\subsubsection{Reasoning-in-a-Haystack}
\label{Reasoning-in-a-Haystack}
We adopt the experimental setup proposed by Fu et al. \cite{fu2024not} to perform the Reasoning-in-a-Haystack evaluation. Unlike the Needle-in-a-Haystack test, this test inserts multiple needles into the haystack, requiring the model to retrieve and reason through them to extract the correct
answer. 
We perform the evaluation on LLaMA2-7B-Chat and Mistral-7B-v0.2-Instruct under a fixed KV cache budget of 50\%. As shown in Table \ref{tab:Reasoning-in-a-Haystack}, Task-KV achieves higher average scores on both models compared to baseline methods. This demonstrates that Task-KV exhibits efficient retrieval and reasoning capabilities across various context length ranges in resource-constrained scenarios.

\begin{figure*}[ht]
    \begin{center}
		\includegraphics[width=0.85\linewidth]{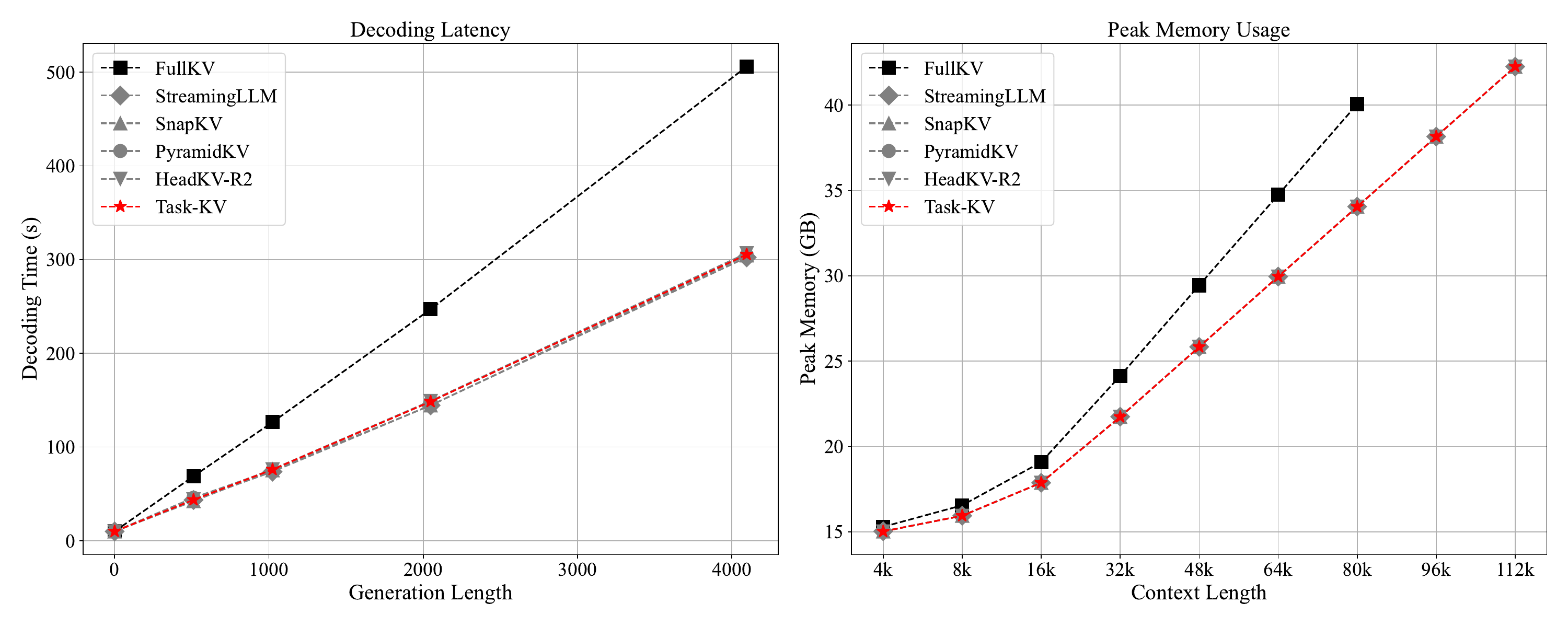}
    \end{center}
    \caption{The Decoding Latency and Peak Memory Usage results.}
    \label{memory_and_time}
\end{figure*}

\subsubsection{Memory and latency}
\label{Memory and latency}
We evaluate the computational efficiency of our Task-KV using the Mistral-7B-Instruct model and set KV cache budget to 40\% for all methods. 
To assess the decoding latency of each method, we use 30K-length data as input and set various generation lengths (1, 512, 1024, 2048, 4096) for comparison.
As shown in the Decoding Latency of Fig. \ref{memory_and_time}, our proposed method achieves the same decoding
latency as other KV cache compression methods. Notably, the decoding time includes both the pre-filling time and the decoding time. Therefore, we can conclude that the pre-filling time for our method and other baselines is almost negligible. 

In addition to decoding latency, we also provide the Peak Memory Usage results, as shown in the
Peak Memory Usage of Fig. \ref{memory_and_time}. Our proposed method achieves performance comparable to other KV cache compression baselines, significantly reducing memory usage compared to the Full KV cache.

\section{Ablation study}
In this section, we first analyze the role of non-heterogeneous attention heads during model inference (Section \ref{Effect of non-heterogeneous heads}). Next, we compare the differences between middle activations and other information compensation methods (Section \ref{Importance of middle activations}). Finally, we provide a detailed explanation of the hyperparameter selection strategy (Section \ref{Hyper parameter selection}). All experiments are conducted using the Mistral-7B-v0.2-Instruct model under a fixed KV cache budget of 50\%.

\begin{figure*}[ht]
    \begin{center}
		\includegraphics[width=0.85\linewidth]{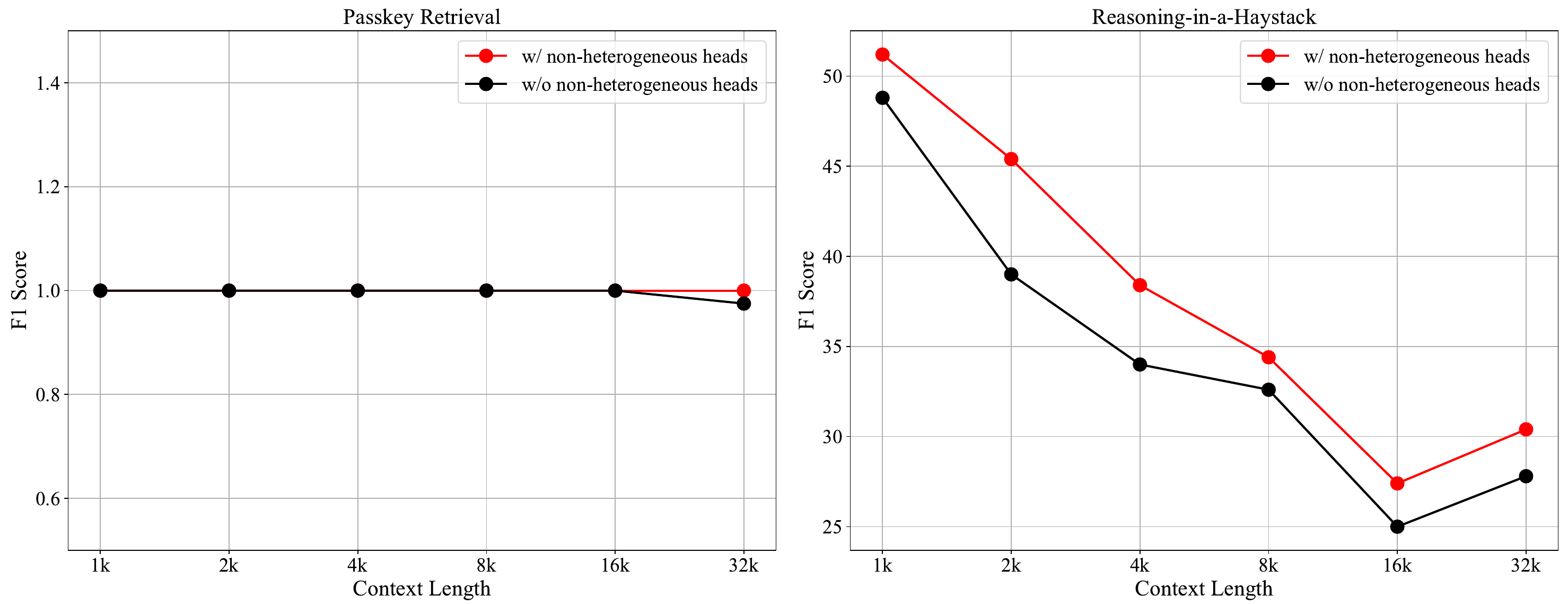}
    \end{center}
    \caption{Results of ablation of non-heterogeneous heads on Passkey Retrieval and Reasoning-in-a-Haystack experiments. }
    \label{ablation_on_non_heter}
\end{figure*}

\subsection{Effect of non-heterogeneous heads}
\label{Effect of non-heterogeneous heads}
To validate the role of non-heterogeneous heads in information reasoning, we conduct two sets of experiments. First, we perform the Passkey Retrieval experiment, designed to evaluate the model’s ability to retrieve random passkeys from long documents, focusing solely on retrieval without involving reasoning. As shown on the left in Fig. \ref{ablation_on_non_heter}, we conduct an ablation study on the non-heterogeneous heads of the top 12 layers of the model. The results indicate that removing the non-heterogeneous heads has minimal impact on the model’s retrieval performance, suggesting that these heads do not contribute to retrieval functionality.

Next, we conduct the Reasoning-in-a-Haystack experiment, which evaluates both retrieval and reasoning capabilities. As illustrated on the right in Fig. \ref{ablation_on_non_heter}, the model’s performance significantly declines when the non-heterogeneous heads are removed. Since retrieval performance is unaffected by the absence of non-heterogeneous heads, this decline can be attributed to weakened reasoning ability. These findings suggest that non-heterogeneous heads play a critical role in enabling information reasoning within the model.

\begin{figure*}[ht]
    \begin{center}
		\includegraphics[width=0.75\linewidth]{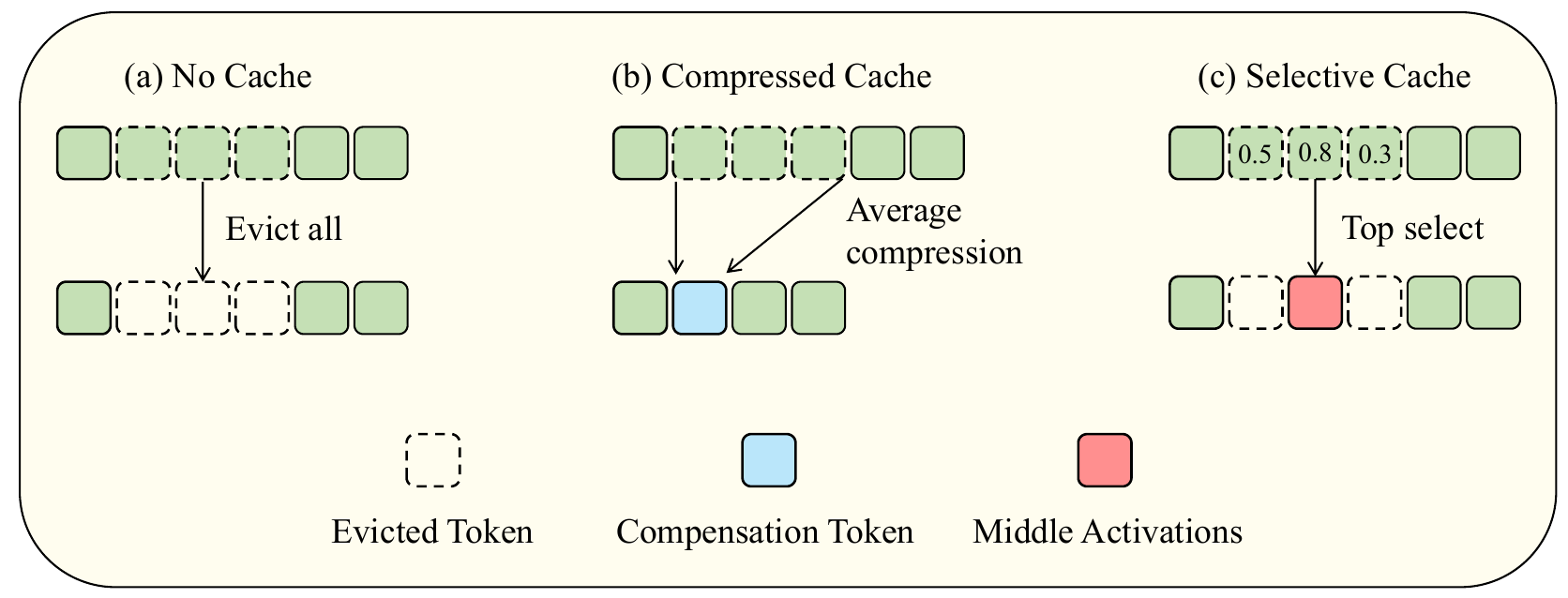}
    \end{center}
    \caption{Differences between the three information compensation methods. (a) No Cache, which retains only sink tokens and recent tokens; (b) Compressed Cache, which averages intermediate tokens into a single compensation token; and (c) Selective Cache, which represents our approach using middle activations.}
    \label{middle_activations}
\end{figure*}

\begin{figure}[ht]
    \begin{center}
		\includegraphics[width=0.95\linewidth]{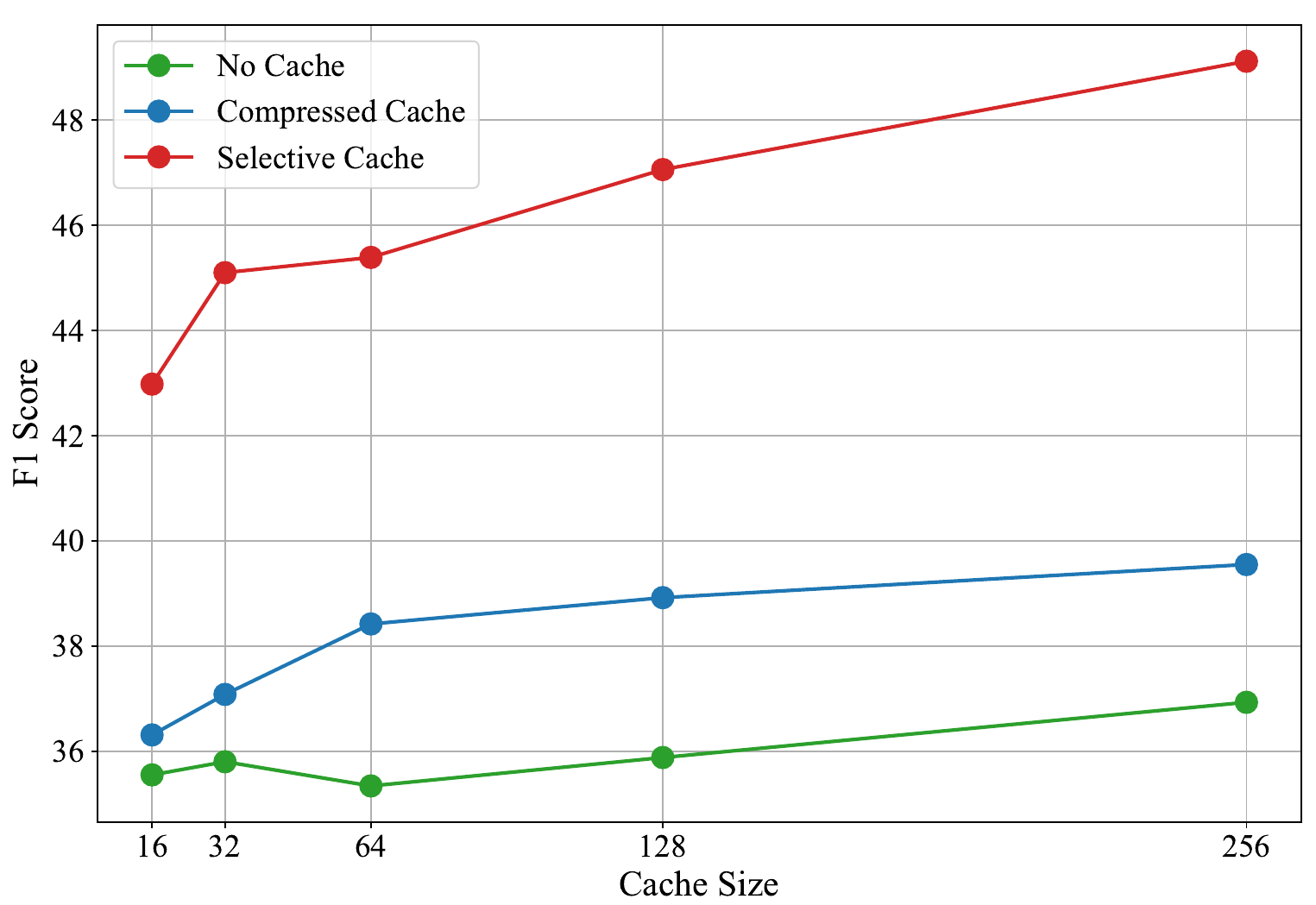}
    \end{center}
    \caption{Results of three information compensation methods on the MultiFieldQA dataset. In the figure, Cache Size represents the number of additional intermediate tokens introduced. To ensure fairness, we keep the number of sink tokens and recent tokens consistent across all methods and observe how different Cache Sizes affect model performance. Specifically, When Cache Size = 16, the No Cache method adds 16 additional tokens to the recent tokens. The Compressed Cache method divides the intermediate tokens into 16 groups, generating one compensation token for each group. The Selective Cache method selects the 16 intermediate tokens with the highest attention scores.}
    \label{importance_middle_activations}
\end{figure}

\subsection{Importance of middle activations }
\label{Importance of middle activations}
We evaluate the impact of three approaches to information compensation on model performance. The distinctions among these methods are illustrated in Fig. \ref{middle_activations}. To ensure fairness, we allocate the same cache size to all three methods and conduct experiments on the MultiFieldQA \cite{bai2023longbench} dataset.

As shown in Fig. \ref{importance_middle_activations}, the results show that incorporating middle activations effectively mitigates information loss and improves the F1 score of the model compared to the other two methods. The No Cache \cite{xiao2024duoattention, xiao2023efficient} method discards all intermediate information, resulting in a significant information gap that cannot be compensated for by increasing the number of recent tokens. The Compressed Cache method \cite{tang2407razorattention}, while partially compensating for information loss by compressing intermediate information into a single token, introduces noise and blurs critical information.
In contrast, our Selective Cache method, which incorporates middle activations, achieves high F1 score even with a small cache size (e.g., cache size = 16). This result indicates that non-heterogeneous heads aggregate key information for reasoning, and this critical information is stored in the middle activations. Retaining these key elements is essential to fully leveraging the reasoning capabilities of non-heterogeneous heads.

\begin{figure*}[t]
    \begin{center}
		\includegraphics[width=0.9\linewidth]{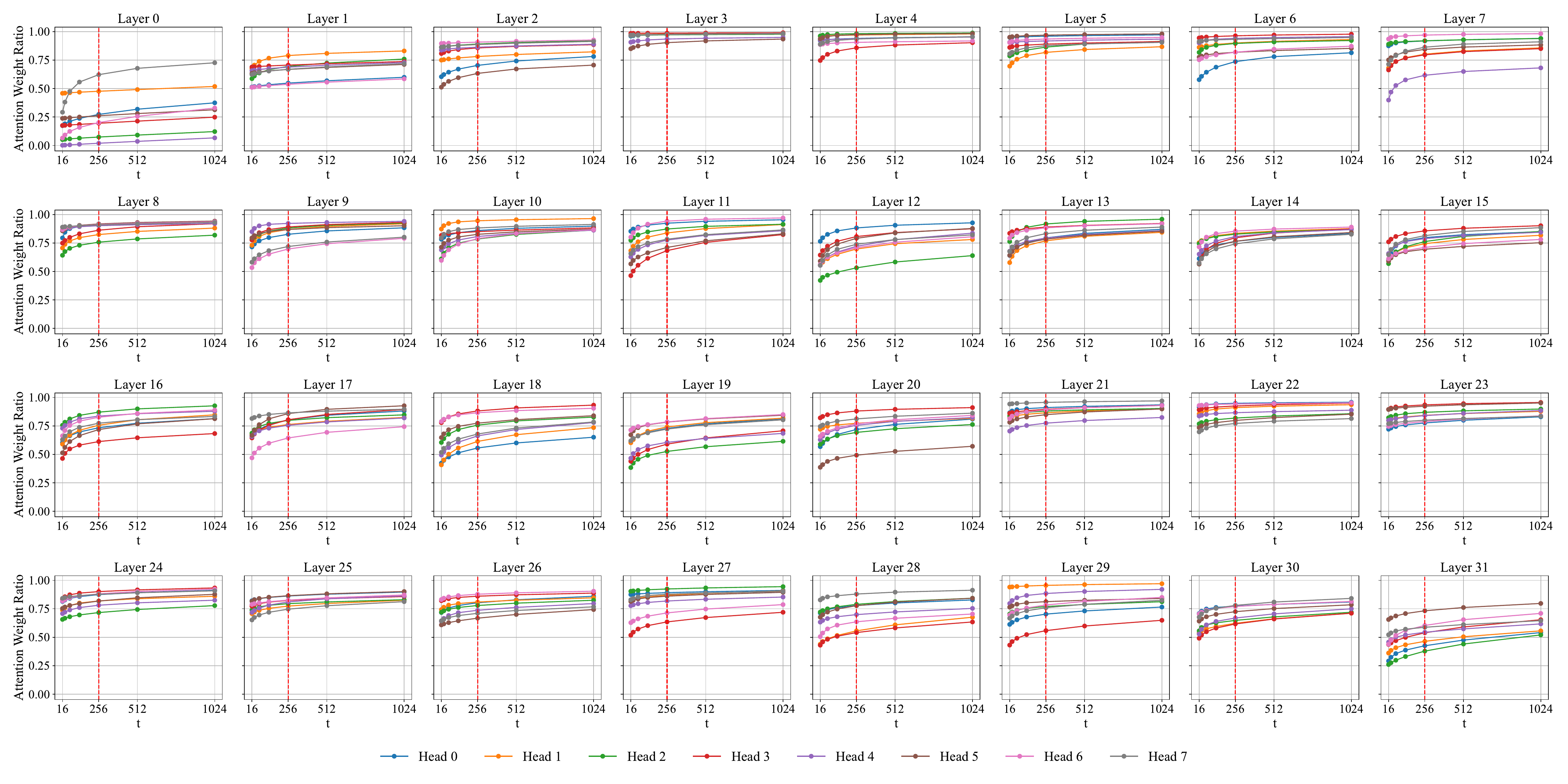}
    \end{center}
    \caption{Analysis of different numbers of top $t$ tokens in attention weights}
    \label{top_t}
\end{figure*}
\begin{figure*}[t]
    \begin{center}
		\includegraphics[width=0.9\linewidth]{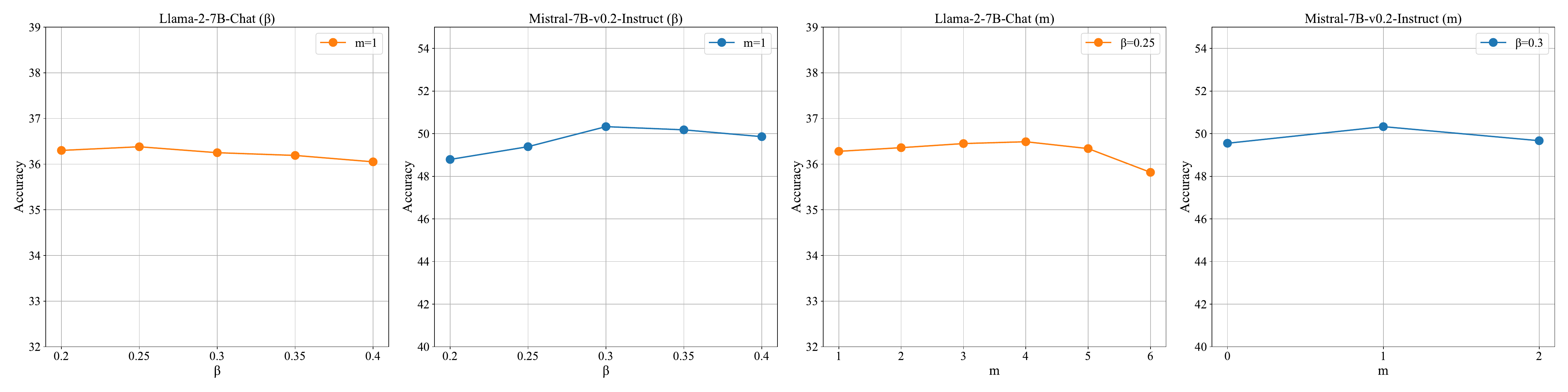}
    \end{center}
    \caption{Ablation experiments on the Llama-2-7B-Chat and Mistral-7B-v0.2-Instruct models for $\beta$ and $m$.}
    \label{ablation_on_beta_m}
\end{figure*}

\subsection{Hyper parameter selection}
\label{Hyper parameter selection}
\subsubsection{Analysis of top $t$ tokens in semantic vector computation}
We use 30k-length data as input and analyze the proportion of top $t$ tokens in attention weights during the computation of semantic vectors. As shown in Fig. \ref{top_t}, even when selecting only a small number of tokens with the highest attention scores (e.g., $t=16$), these tokens still account for a substantial proportion of the attention weights (e.g., Layer 3). The overall trend indicates that when $t=256$, the proportion reaches a turning point. Beyond this threshold, the rate of increase in the weight proportion progressively slows as $t$ increases further. Consequently, we set $t=256$ for semantic vector computation to effectively balance computational cost with accuracy requirements.

\subsubsection{Analysis of $\beta$ and $m$ in layer-wise heterogeneous heads allocation}
\label{Analysis of beta and m }
When the values of $\beta$ and $ m $ are set too high, the KV cache budget for non-heterogeneous heads is compressed, which negatively impacts the model's reasoning capability. Conversely, if these values are too low, the model may fail to fully understand the task semantics. Therefore, we first fix \( m=1 \), and then select different values of \( \beta \) from the set \{0.2, 0.25, 0.3, 0.35, 0.4\} to observe the changes in model F1 score on the MultiFieldQA \cite{bai2023longbench} dataset. As shown in the left two figures in Fig. \ref{ablation_on_beta_m}, when \( \beta=0.25 \) and \( \beta=0.3 \), Task-KV achieves the highest score on the Llama-2-7B-Chat and Mistral-7B-v0.2-Instruct models, respectively. For the Llama model, we fix \( \beta=0.25 \) and test the results for \( m \) values in \{1, 2, 3, 4, 5, 6\}, while for the Mistral model, we fix \( \beta=0.3 \) and test the results for \( m \) values in \{0, 1, 2\}. As shown in the right two figures in Fig. \ref{ablation_on_beta_m}, when \( m=4 \) and \( m=1 \), Task-KV demonstrates optimal performance on two models, respectively. Furthermore, Fig. \ref{ablation_on_beta_m} show that the score remains relatively stable in the middle range, while larger fluctuations occur at the extremes. This suggests that as long as the number of heterogeneous heads is within a reasonable range, it has little impact on the final performance of the model and is robust to changes in parameters. However, when the parameters fall within extreme ranges, it may lead to performance instability.

\section{Conclusion}
In this study, we theoretically and experimentally demonstrate the significance of heterogeneous heads for model's outputs. Heterogeneous heads capture semantic information from diverse perspectives, which helps enhance the model's representational and generalization abilities. Furthermore, our experiments confirm that the heterogeneous heads activated by different types of tasks exhibit significant differences. Based on these insights, we propose a novel KV cache compression method, called Task-KV, which dynamically allocates KV cache budgets by leveraging task-aware semantic differences among attention heads. Task-KV consists of two key components: a semantic separator and a KV cache allocation strategy. The semantic separator efficiently computes the semantic vectors of attention heads through a two-stage optimization process, and selects task-relevant heterogeneous and non-heterogeneous heads based on the distance from the semantic center. The KV cache allocation strategy assigns the full KV cache budget to the heterogeneous heads, ensuring the completeness of multi-perspective semantic information. For non-heterogeneous heads, our extensive experiments confirm their critical role in information aggregation and reasoning. Consequently, we allocate a small number of sink tokens and recent tokens to non-heterogeneous heads to maintain their basic reasoning capabilities, while introducing middle activations to retain crucial aggregated information. We comprehensively evaluate Task-KV across multiple benchmarks, models, and long-context tasks. The overall results demonstrate that our method achieves superior performance while maintaining computational efficiency.

\bibliographystyle{IEEEtran}

\bibliography{ref}

\end{document}